\documentclass[10pt, journal]{IEEEtran}

\usepackage{pifont}

\usepackage{amsmath,amssymb,amsfonts}
\usepackage{bm} 

\usepackage{graphicx}
\usepackage{tikz}

\usepackage{booktabs}
\usepackage{multirow}
\usepackage{makecell}
\usepackage[table,xcdraw]{xcolor}

\usepackage{algorithmic}
\usepackage{algorithm}
\usepackage{enumitem}

\usepackage{cite}
\usepackage{url}
\usepackage{microtype}
\usepackage{balance} 

\usepackage[font=small,labelfont=bf]{caption}
\usepackage{orcidlink}

\title{MAPRPose: Mask-Aware Proposal and Amodal Refinement for Multi-Object 6D Pose Estimation}

\author{Yang~Luo\orcidlink{0009-0002-7124-2296}, 
        Yan~Gong\orcidlink{0000-0002-3148-8286}, 
        Yongsheng~Gao*\orcidlink{0000-0002-1555-8328}, 
        Xiaoying~Sun*\orcidlink{0000-0001-9672-8554}, 
        Jie~Zhao\orcidlink{0000-0002-6086-9387},~\IEEEmembership{Senior Member,~IEEE}

\thanks{Manuscript received April 21, 2026. This work was supported in part by the National Science and Technology Major Project of China under Grant 2025ZD1603200, in part by the National Science Fund for Outstanding Young Scholars of the National Natural Science Foundation of China under Grant 52025054, and in part by the Natural Science Foundation of Heilongjiang Province of China under Grant LH2021E076. (\textit{Corresponding authors: Yongsheng Gao; Xiaoying Sun.})}

\thanks{Yang Luo, Yan Gong, Yongsheng Gao, and Jie Zhao are with the State Key Laboratory of Robotics and Systems, Harbin Institute of Technology, Harbin 150001, China (e-mail: christoluo@outlook.com; gongyan2020@foxmail.com; gaoys@hit.edu.cn; jzhao@hit.edu.cn).}%

\thanks{Xiaoying Sun is with the School of Civil Engineering, Harbin Institute of Technology, Harbin 150001, China (e-mail: sxy\_hit@163.com).}%

}

\begin{document}
\maketitle

\begin{abstract}
6D object pose estimation in cluttered scenes remains challenging due to severe occlusion and sensor noise. We propose \textbf{MAPRPose}, a two-stage framework that leverages mask-aware correspondences for pose proposal and amodal-driven Region-of-Interest (ROI) prediction for robust refinement.  In the \textbf{Mask-Aware Pose Proposal (MAPP)} stage, we lift 2D correspondences into 3D space to establish reliable keypoint matches and generate geometrically consistent pose hypotheses based on correspondence-level scoring, from which the top-$K$ candidates are selected. In the refinement stage, we introduce a tensorized render-and-compare pipeline integrated with an \textbf{Amodal Mask Prediction and ROI Re-Alignment (AMPR)} module. By reconstructing complete object geometry and dynamically adjusting the ROI, AMPR mitigates localization errors and spatial misalignment under heavy occlusion. Furthermore, our GPU-accelerated RGB-XYZ reprojection enables simultaneous refinement of all $N \times B$ pose hypotheses in a single forward pass. 
\end{abstract}


\section{Introduction}
6D object pose estimation, which recovers the 3D rotation and translation of objects from visual observations, is fundamental for autonomous robotic systems~\cite{depth_rgb_fusion2019, 2021survey, Zhou2022, ConfidencePose2022, HFF6DPose2022, Yuanwei2024, TG-Pose2024, Tu2025, Any6D2025, liu2026survey, xia2025vlm6d, liu2025activepose, Cheng2025, ponimatkin2025freepose, yang2026active, zhang2026enhanced, GCM-Pose2026}. A particularly important yet challenging scenario is zero-shot pose estimation of novel objects, given their CAD models, under heavy occlusion and background clutter with strict requirements on computational efficiency. This capability is critical for applications such as conveyor-belt manipulation and large-scale warehouse bin-picking in unstructured environments~\cite{liang2025dynamicpose}. Although supervised methods excel at instance-level estimation for known objects, real-world deployment requires robust generalization to novel instances. This challenge is further compounded in cluttered environments with partial observations. Moreover, practical systems demand high computational efficiency for real-world deployment.

To mitigate the adverse effects of occlusion and background clutter, recent studies have increasingly incorporated mask-based strategies into the pose estimation pipeline. For instance, methods such as PoseCNN~\cite{PoseCNN2018} and One2Any~\cite{one2any2025} employ instance-level masks as spatial priors, whereas multi-task architectures like SO-Pose~\cite{SO-Pose} and Self6D~\cite{Self6D} integrate amodal segmentation to recover complete object geometry. Despite these advancements, such approaches still suffer from two critical limitations. First, their heavy reliance on object-centric cropping frequently yields suboptimal initial translation estimates under severe occlusion. Second, this rigid, sequential optimization process makes them highly susceptible to local optima, often requiring dozens of iterations to achieve high pose accuracy, which severely limits their scalability in complex, multi-object scenes.

\begin{figure}
     \centering
    \includegraphics[trim={6.5cm 2.8cm 10cm 2.6cm}, clip, width=1\linewidth]{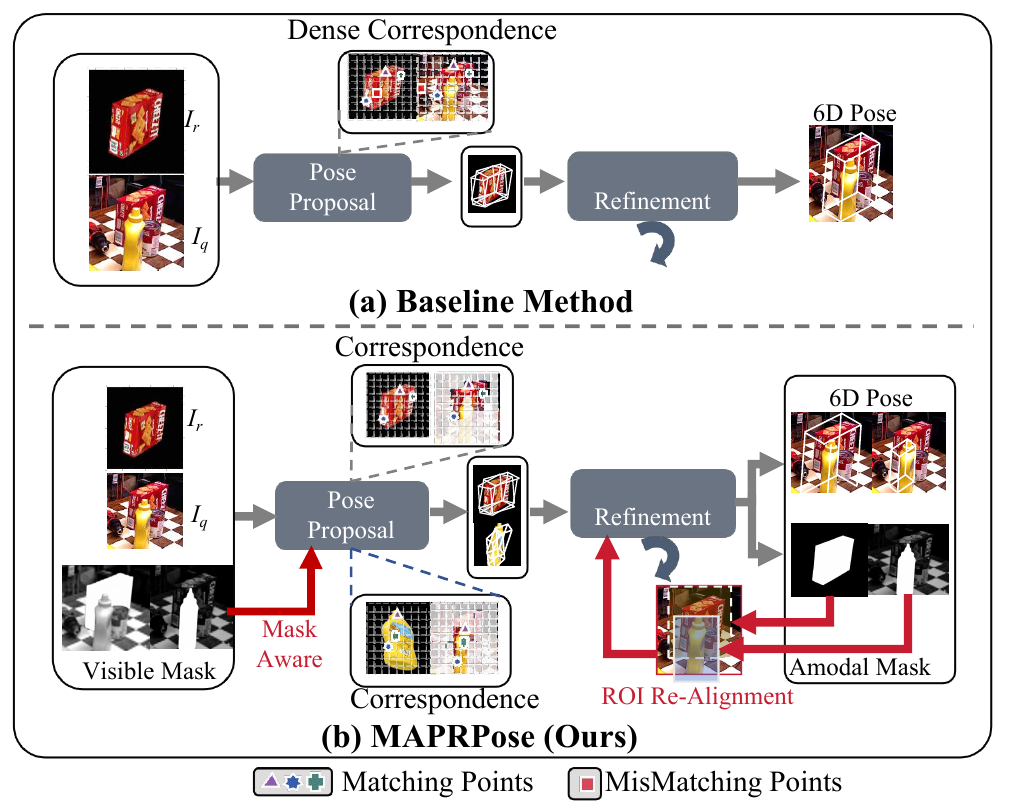}
    \captionsetup{
        font=small,
        labelfont=bf,
        justification=justified,
    }
\caption{\small\textbf{Comparison of MAPRPose with Baseline.} 
\textbf{(a) Baseline} methods are typically limited by (1): indiscriminate matching of foreground and background, (2): mask-free alignment, and (3): inefficient single-object processing. 
\textbf{(b) MAPRPose} employs (1) mask-aware coarse matching and (2) parallel multi-object processing for efficient inference, while integrating (3) amodal-driven ROI re-alignment to improve pose accuracy.}
\label{fig:comparison}
\end{figure}

To overcome the susceptibility to local optima, recent works~\cite{MegaPose, FoundationPose, Gigapose} have shifted toward the render-and-compare paradigm. By iteratively rendering pose hypotheses and comparing them against the query image, methods such as MegaPose~\cite{MegaPose} and FoundationPose~\cite{FoundationPose} achieve strong robustness and zero-shot generalization. However, this paradigm faces a fundamental trade-off between hypothesis coverage and computational efficiency in complex, occluded, and multi-object scenes. Specifically, evaluating a large number of hypotheses per inference introduces substantial computational overhead. Although some hybrid approaches~\cite{Co-op, Gigapose} use coarse-to-fine selection to prune the search space, they often lack explicit geometric grounding and remain sensitive to extreme clutter. Moreover, their reliance on sequential processing for each object and hypothesis further limits overall efficiency and scalability in cluttered environments.

These limitations indicate that the key bottleneck lies in the inefficient representation and evaluation of the hypothesis space. This motivates our proposed framework, MAPRPose, which leverages mask-aware reasoning to generate high-quality proposals and enables tensorized parallel inference for efficient, amodal-driven refinement for multiple objects.

As illustrated in Fig.~\ref{fig:comparison}, we depart from baseline frameworks~\cite{Co-op, Gigapose} that rely on unconstrained feature matching. Instead, we formulate the 6D pose estimation as a partial-to-amodal geometric alignment problem. We present MAPRPose, a unified framework that synergizes mask-aware proposal generation and amodal-driven refinement to address the decoupled components of 6D pose initialization and optimization. In the pose proposal stage, we introduce Mask-Aware Pose Proposal (MAPP) to provide rotation initialization. By constraining correspondence search within the visible object manifold, MAPP neutralizes background clutter and prunes the search space from hundreds of templates to a compact set of Top-$K$ hypotheses. This not only avoids the local optima caused by the initialization drift but also reduces computational cost. To alleviate initialization bias in translation estimation, we introduce the Amodal Mask Prediction and ROI Re-Alignment (AMPR) mechanism. While conventional pipelines frequently suffer from suboptimal translation estimates due to cropping biased towards visible regions, AMPR reconstructs the full object geometry to calibrate the ROI's center and scale prior to refinement. This ensures that the subsequent optimization is anchored to the true object centroid, effectively rectifying the localization errors inherent in partial observations. Crucially, our framework re-architects the render-and-compare paradigm into a fully tensorized refinement and re-projection pipeline. Conventional approaches typically process hypotheses and objects sequentially.

The technical contributions are summarized as follows:
\begin{itemize}[topsep=0pt, leftmargin=*]
\item We propose \textbf{MAPP}, which generates robust rotation initialization by distilling mask-aware correspondences into a compact set of hypotheses, reducing background interference and search latency.
\item We introduce \textbf{AMPR}, which corrects translation biases by inferring amodal object geometry, enabling a more accurately centered ROI for  refinement.

\end{itemize}

\section{Related Work}
\subsection{CAD Model-based Methods}

CAD-model-based methods are the predominant approach for zero-shot 6D pose estimation of novel objects. These methods can be broadly categorized into correspondence-based methods and render \& compare methods, with many recent works~\cite{Gigapose, MegaPose, Co-op, FreeZeV2} utilizing hybrid two-stage pipelines that combine fast proposal generation with iterative refinement. Correspondence-based methods, such as DenseFusion~\cite{wang2019densefusion}, DPOD~\cite{DPOD}, SurfEmb~\cite{haugaard2022surfemb}, GeoPose~\cite{GeoPose2022} and others~\cite{GCPose2024, FoundPose, Pos3R2025, NOCS}, establish dense 2D-3D correspondences or establish matching relationships via iterative feature fusion between query observations and CAD-derived representations. These established correspondences are then processed by geometric solvers (e.g., PnP~\cite{lepetit2009epnp} or RANSAC~\cite{fischler1981ransac}) for final pose estimation. Although these methods are computationally efficient, they remain vulnerable to segmentation inaccuracies. In particular, severe occlusions often lead to a significant initialization drift in pose estimation. Moreover, the lack of discriminative local geometric cues makes it difficult to distinguish foreground from background, resulting in erroneous correspondences and preventing accurate 6D pose recovery from partially visible observations.

Render \& compare methods, including MegaPose~\cite{MegaPose}, FoundationPose~\cite{FoundationPose}, SAM-6D~\cite{SAM-6D}, FreeZe~\cite{FreeZe}, FreeZeV2~\cite{FreeZeV2}, and Geo6D~\cite{huang2024matchu}, evaluate multiple pose hypotheses by reducing the visual or geometric discrepancy between the rendered model and the observed query. Although these frameworks are generally more robust to occlusion and environmental noise than pure correspondence-based techniques, they often incur higher computational overhead due to the iterative nature of the alignment process. To further enhance this paradigm, specialized approaches have been developed for complex scenarios: EPOS~\cite{EPOS} introduces effective handling of both discrete and continuous symmetries, while CosyPose~\cite{labbe2020cosypose} enforces global geometric consistency by optimizing 6D poses across multiple views.

Despite their differences, these paradigms share two critical limitations. First, hybrid pipelines that combine both strategies (e.g., Co-op~\cite{Co-op}, GigaPose~\cite{Gigapose}, FreeZeV2~\cite{FreeZeV2}) frequently suffer from dense correspondence matching~\cite{Co-op, Gigapose} that is prone to corruption by cluttered backgrounds and severe occlusion. This lack of explicit geometric grounding during feature matching leads to decoupled reasoning, making it difficult to achieve robust pose estimation. Second, the computational scalability of these methods is strictly limited by the heavy per-object rendering overhead (e.g., nvdiffrast~\cite{nvdiffrast}), which hinders efficient parallel computing in multi-object scenes. These challenges highlight the need for approaches that integrate mask-aware correspondence modeling with scalable geometric reasoning across both proposal and refinement stages, ensuring that optimization remains focused on the object's visible manifold.

\subsection{Occlusion-Aware Pose Estimation}
Occlusion-aware pose estimation has evolved from correspondence-based methods~\cite{PVNet2019, ONDA-Pose, GDR-Net, wang2024oapose} to mask-guided frameworks like~\cite{SO-Pose}, ~\cite{Mask6D}, ~\cite{liu2025occlusionaware}, which leverage visible masks as spatial priors, but remain sensitive to incomplete observations. However, a fundamental bottleneck persists: the systemic spatial-scale misalignment inherent in traditional, visible-mask-driven Region of Interest (ROI) extraction. Existing pipelines typically operate in an open-loop manner, where cropping based solely on visible fragments shifts the image-plane localization ($t_x, t_y$) and introduces critical scale-inconsistency during resizing. This artificial magnification distorts the perceived object scale relative to its canonical CAD reference, inevitably corrupting depth ($t_z$) inference. 

To address these gaps, recent trends have shifted to amodal-aware geometric recovery. For instance, UA-Pose~\cite{li2025uapose} introduces uncertainty-aware completion to handle partial references, while other works~\cite{Amodal3R2025, Oryon2024} explore amodal completion and open-vocabulary reasoning to resolve pose ambiguities. Building upon this paradigm, we propose a multi-hypothesis framework that replaces the conventional one-pass cropping with a recursive feedback mechanism. By dynamically re-aligning the ROI based on predicted full-object geometry, our approach maintains global geometric consistency and rectifies the scale-inconsistency that plagues previous occlusion-aware pipelines.
\begin{figure*}[t]
    \centering
    \includegraphics[trim={6.1cm 0.3cm 3.6cm 0.1cm}, clip, width=1\linewidth]{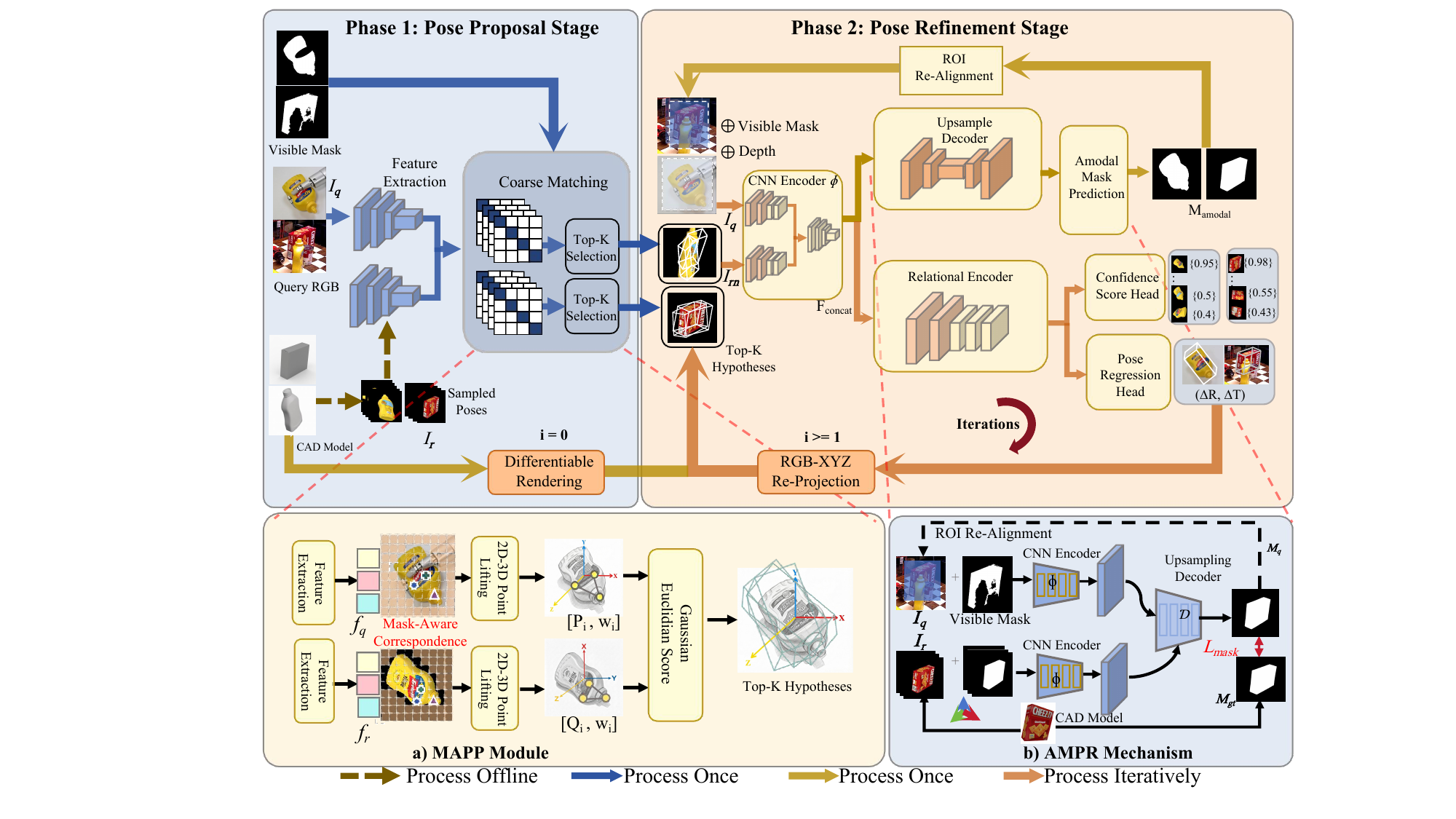}
    \captionsetup{
        font=small,
        labelfont=bf,
        justification=justified,
    }
    \caption{\small\textbf{Overall Architecture of MAPRPose.} Our framework follows a coarse-to-fine paradigm consisting of two integrated stages. \textbf{Phase 1: MAPP.} Visible masks are utilized to constrain patch-level matching between the query image and multi-view CAD-rendered templates. These mask-aware correspondences are lifted to 3D keypoints to generate a compact set of geometrically consistent pose hypotheses. \textbf{Phase 2: Pose Refinement.} These hypotheses are refined via a network that \textbf{embeds} the \textbf{AMPR} mechanism. Specifically, the architecture is designed for \textbf{parallel processing of multiple object instances and hypotheses} in a single forward pass. The \textbf{AMPR} branch reconstructs the full amodal shape to \textbf{re-center and normalize the scale} of target objects, established via a feedback loop. To maximize computational throughput, we develop a \textbf{hybrid refinement strategy}: high-fidelity CAD rendering is performed only for initialization ($i=0$), while subsequent iterations ($i>0$) leverage a \textbf{RGB-XYZ re-projection}.} 
    \label{fig:Overview of Framework}
\end{figure*}

\section{Methodology}
\subsection{Overview}
We now present MAPRPose, a two-stage framework that combines mask-aware pose proposal with amodal-driven refinement (Fig.~\ref{fig:Overview of Framework}). In the MAPP stage, correspondences between the target query and reference templates are extracted using DINOv2~\cite{DINOv2} features and lifted to 3D to generate top-$K$ geometrically consistent pose hypotheses. These proposals are then passed to the Pose Refinement stage, which incorporates an AMPR module to rectify location and scale discrepancies for precise spatial consistency. Furthermore, the refinement stage predicts relative pose increments and performs fast RGB-XYZ re-projection to synthesize pixel-aligned reference-to-query representations.

\subsection{Mask-Aware Pose Proposal Stage}

The Mask-Aware Pose Proposal (MAPP) stage aims to generate a compact yet expressive set of pose hypotheses by approximating the optimal solution of a constrained correspondence-based alignment problem. 
Given a query image $I_q$ and an object CAD model $\mathcal{M}$, we seek an initial pose $\mathbf{T} \in SE(3)$ that maximizes geometric consistency under partial observations. 
Instead of performing exhaustive search over $SO(3)$, we construct a geometry-aware hypothesis space that incorporates visibility priors to restrict sampling to physically plausible configurations.

\begin{equation}
\mathcal{R} = \left\{ \mathbf{R} \in \mathcal{R}_{\text{raw}} \;\middle|\; \mathbf{e}_z^\top \mathbf{t}(\mathbf{R}) \geq \sigma D \right\},
\end{equation}
where $D$ is the object diameter and $\sigma$ controls the admissible viewing range. This constraint removes viewpoints that are unlikely to yield observable projections. 
For each valid rotation, we further sample in-plane rotations $\theta$ with step size $\Delta \theta$, forming a structured hypothesis set:
\begin{equation}
\mathcal{G}_{\text{rot}} = \left\{ \mathbf{R} \cdot \mathbf{R}_z(\theta) \;\middle|\; \mathbf{R} \in \mathcal{R},\; \theta \in \Theta \right\}.
\end{equation}
This construction can be viewed as imposing a prior over $SO(3)$ biased toward observable configurations. 
In addition, we adopt a canonicalized depth setting by representing the camera translation along the optical axis as a fixed translation variable $\mathbf{T}_z$, where $\mathbf{T}_z \in \mathbb{R}^3$ denotes a canonical translation aligned with the viewing direction and proportional to the object scale. Specifically, we define $\mathbf{T}_z = \alpha D \mathbf{e}_z$, where $D$ denotes the object diameter, $\mathbf{e}_z$ is the optical axis direction, and $\alpha$ is a constant controlling the effective viewing distance (set to $10$ in our implementation). This design reduces the sensitivity of the rendered appearance to absolute depth variations, as moderate changes along the $z$-axis induce only second-order effects on the projected geometry under perspective projection, while preserving the relative silhouette and texture consistency used for matching.

For each initial pose $\mathbf[\mathbf{R}| \mathbf{T}_z] \in SO(3)$, we retrieve a pre-rendered template image along with its dense features $\mathbf{f}_r(j) \in \mathbb{R}^C$, which are precomputed offline for all sampled viewpoints to avoid redundant computation. 
The query features $\mathbf{f}_q(i) \in \mathbb{R}^C$ are extracted on-the-fly from $I_q$. 
To suppress background interference, we introduce a mask-aware similarity kernel:
\begin{equation}
S(i,j) = \left( m_q(i)\, m_r(j) \right) \cdot \langle \mathbf{f}_q(i), \mathbf{f}_r(j) \rangle,
\end{equation}
where $m_q(i), m_r(j) \in \{0,1\}$ denote binary visibility masks. 
This formulation induces a structured sparsity over the correspondence space by restricting matching to foreground regions, while preserving feature magnitude allows highly discriminative patches to dominate the matching process. 
Dense correspondences are established via nearest-neighbor assignment $j_i = \arg\max_j S(i,j)$.

Each correspondence $(i, j_i)$ is then lifted to 3D point pairs $(\mathbf{Q}_i, \mathbf{P}_i)$ using depth back-projection and rendered geometry, respectively. 
We evaluate alignment quality using a translation-invariant geometric consistency score. 
Let $w_i = S(i, j_i)$ and define weighted centroids:
\begin{equation}
\bar{\mathbf{P}} = \frac{\sum_i w_i \mathbf{P}_i}{\sum_i w_i}, 
\quad
\bar{\mathbf{Q}} = \frac{\sum_i w_i \mathbf{Q}_i}{\sum_i w_i}.
\end{equation}
The final scoring function is given by:
\begin{equation}
\mathcal{E}(\mathbf{T}) = \sum_i w_i \exp\left(
- \frac{
\left\| (\mathbf{P}_i - \bar{\mathbf{P}}) - (\mathbf{Q}_i - \bar{\mathbf{Q}}) \right\|_2^2
}{2\sigma^2}
\right),
\end{equation}
which can be interpreted as a Gaussian kernel density estimate over alignment residuals, measuring the rigidity of the induced correspondence field. 
We retain the top-$K$ hypotheses with the highest scores and select the best rotation $\mathbf{R}^*$ accordingly.

Finally, translation is estimated independently to decouple rotational ambiguity. 
We compute the median pixel location within the object mask along with its corresponding depth value, and back-project it into 3D space using camera intrinsics:
\begin{equation}
\mathbf{t}^* = z \mathbf{K}^{-1} [u, v, 1]^\top.
\end{equation}
The resulting initialization $\mathbf{T}_{\text{init}} = (\mathbf{R}^*, \mathbf{t}^*)$ provides a robust starting point for subsequent refinement, while avoiding bias introduced by asymmetric geometry or partial visibility.

\bibliographystyle{IEEEtran}
\bibliography{refs}

\end{document}